\documentclass[10pt,twocolumn,letterpaper]{article}

\usepackage[pagenumbers]{cvpr} 

\usepackage[table]{xcolor}
\usepackage{multirow}

\definecolor{cvprblue}{rgb}{0.21,0.49,0.74}
\usepackage[pagebackref,breaklinks,colorlinks,citecolor=cvprblue]{hyperref}

\def\paperID{****} 
\def\confName{CVPR}
\def\confYear{2024}

\title{VLM-Eval: A General Evaluation on Video Large Language Models}

\author{
Shuailin Li$^{1}$\footnotemark[1], Yuang Zhang$^{2}$\footnotemark[1] \footnotemark[2], Yucheng Zhao$^{1}$\footnotemark[1], Qiuyue Wang$^{1}$, Fan Jia$^{1}$, Yingfei Liu$^{1}$, Tiancai Wang$^{1}$ \footnotemark[3] \\
$^{1}$MEGVII Technology \quad $^{2}$Shanghai Jiao Tong University \\
}

\begin{document}
\maketitle

\renewcommand{\thefootnote}{\fnsymbol{footnote}}
\footnotetext[1]{Equal contribution.}
\footnotetext[2]{This work was done during the internship at MEGVII Technology.}
\footnotetext[3]{Corresponding author.}
\renewcommand{\thefootnote}{\arabic{footnote}}

\begin{abstract}
  Despite the rapid development of video Large Language Models (LLMs), a comprehensive evaluation is still absent.
  In this paper, we introduce a unified evaluation that encompasses multiple video tasks, including captioning, question and answering, retrieval, and action recognition. In addition to conventional metrics, we showcase how GPT-based evaluation can match human-like performance in assessing response quality across multiple aspects.
  We propose a simple baseline: Video-LLaVA, which uses a single linear projection and outperforms existing video LLMs.
  Finally, we evaluate video LLMs beyond academic datasets, which show encouraging recognition and reasoning capabilities in driving scenarios with only hundreds of video-instruction pairs for fine-tuning.
  We hope our work can serve as a unified evaluation for video LLMs, and help expand more practical scenarios.
  The evaluation code will be available soon.
\end{abstract}


\section{Introduction}
\label{sec:intro}

Video understanding is pivotal to real-world applications, including embodied robotic agents, disability services, and autonomous driving.
Previous paradigms mainly adopt pre-trained foundation models~\cite{radford2021learning,tong2022videomae,wang2022internvideo} and finetune them for specific tasks. They require extensive data annotation and hand-crafted strategies, which limit their adaptability to open-ended applications.
Recent efforts in connecting video and Large Language Models (LLMs)~\cite{OpenAI2023, chiang2023vicuna, touvron2023llama, taori2023stanford} have significantly enhanced general video understanding in zero-shot settings.
Without specific training, current video LLMs~\cite{zhang2023video,li2023videochat,li2023blip} can interact with humans through a natural language interface and perform spatial-temporal perception, reasoning, and causal inference tasks.

\begin{figure}
  \centering
  \includegraphics[width=\linewidth]{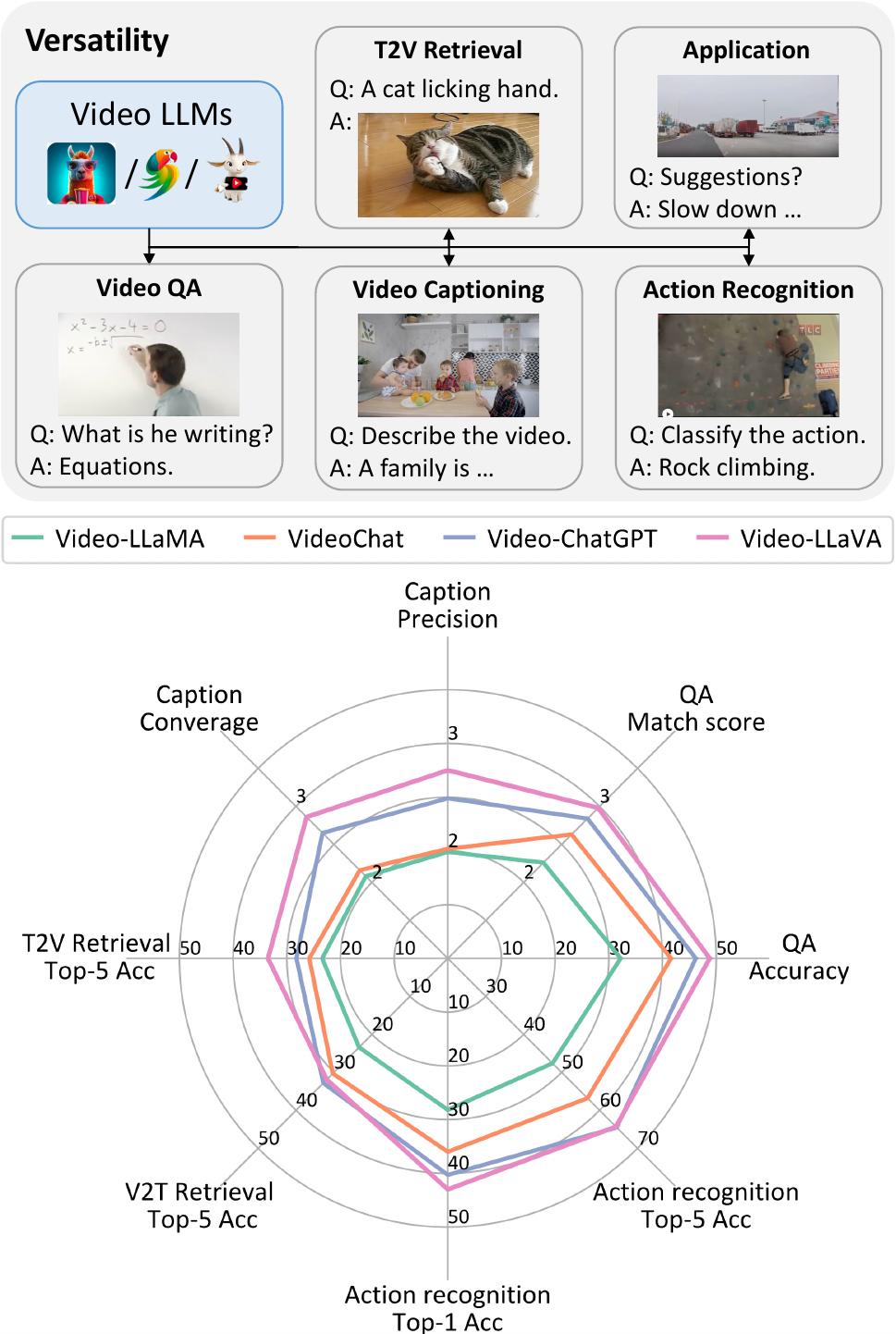}
  \caption{Evaluation of video Large Language Models (LLMs): a multidimensional study of their video understanding capabilities.}
  \label{fig:fig1}
\end{figure}

\begin{figure*}
  \centering
  \includegraphics[width=\linewidth]{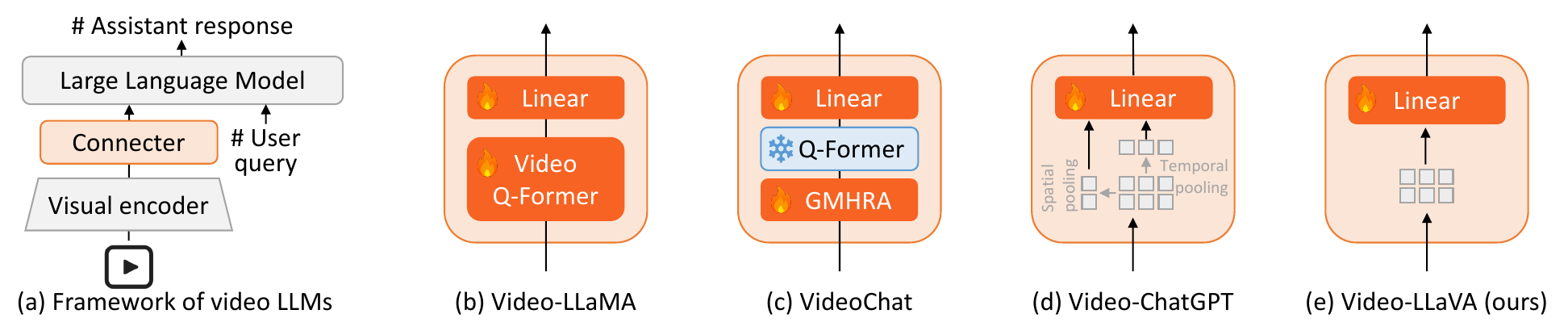}
  \caption{Comparison of Video LLMs. The snow icon denotes frozen parameters and the fire icon indicates parameters tuned in training.}
  \label{fig:cmp}
\end{figure*}

However, evaluating video LLMs is a significant challenge. It requires evaluating open-ended responses and considering relevance to video content and user prompts.
A high-quality response should meet multiple criteria. First, it should be comprehensive and cover all aspects of the user's query while fully reflecting video content. Second, the response must be precise and grounded on video content and user prompts without any hallucination. In addition, it should be focused on addressing user prompts directly without generating excessive or irrelevant responses.

Existing research~\cite{maaz2023video,zhang2023video,li2023videochat} primarily focuses on qualitative evaluation, resulting in a lack of objectivity, comprehensiveness, and automation.
In this paper, we propose a thorough evaluation that covers GPT-based, retrieval-based, and conventional metrics across various tasks and datasets.
To tackle the challenge of evaluating open-ended conversations, we evaluate and incorporate ChatGPT \cite{OpenAI2023} as a quality assessment agent.
In contrast to previous efforts using GPT-based scoring, our focus is on the validity of GPT-based metrics for question-answer and video captioning tasks. We evaluate video LLMs based on their response comprehensiveness, correctness, and conciseness. More importantly, we chose the criteria that GPT scores are consistent with human scores.
After validating ChatGPT's ability to evaluate video LLMs, we relieve the human burden with ChatGPT.
Our evaluation, summarized in Fig.~\ref{fig:fig1}, aims to serve as a groundwork for future study and facilitate a deeper understanding of existing video LLMs.

To further explore the impact of video-to-text connector between visual encoder and LLM, we propose a simple video LLM baseline following LLaVA~\cite{liu2023visual}. The proposed model is named Video-LLaVA, where we directly feed multi-frame features into the LLM without Q-former~\cite{li2023blip} or spatial/temporal pooling~\cite{maaz2023video}. The proposed baseline outperforms prior methods in numerous video-related tasks, indicating that connecting video features to LLMs is of essence, while the design of adapters is less significant.

Finally, we look beyond academic datasets to see how they apply to specific industries. We present a case study in driving scenarios to better understand the capabilities of video LLMs. Our research focuses on investigating the few-shot capability of video LLMs through supervised fine-tuning. We collect hundreds of video clips of roads and then annotate them with detailed captions such as vehicle location, traffic signs, causes of traffic accidents, and driving advice. These make up the video-instruction pairs for the supervised fine-tuning phase. Using such a small dataset, our model demonstrates perception, understanding, reasoning, and planning capabilities in traffic scenarios. This suggests that video LLM is a promising path to autonomous driving.

In summary, our contributions are as follows,
\begin{itemize}
  \item We conducted a comprehensive evaluation of video LLMs, verifying the effectiveness of the ChatGPT score, while also using retrieval-based and conventional metrics.
  \item We build Video-LLaVA as a baseline archiving SoTA performance to show that a simple connector can work well.
  \item We demonstrate the effectiveness of video LLMs in a specific industrial scenario beyond academic datasets.
\end{itemize}


\section{Related Work}

\subsection{LLMs and Multimodal LLMs}
\noindent \textbf{LLMs} Large language models~\cite{radford2019language,devlin2018bert,raffel2020exploring,brown2020language,ouyang2022training,OpenAI2023,zhang2022opt,touvron2023llama,zeng2022glm,taori2023stanford,chiang2023vicuna} have gained significant interest in many natural language processing (NLP) tasks, featuring extraordinary performance and adaptability.
LLMs excel at textual understanding, generation, and reasoning capabilities through large-scale pre-training, and show exceptional zero-shot and emergent capabilities~\cite{wei2022emergent} when scaling up model size.

\noindent \textbf{Image LLMs}
Besides NLP tasks, many researchers leverage LLMs for general image understanding.
Some works~\cite{liang2023taskmatrix, shen2023hugginggpt, wu2023visual} employ the detection models to provide the perception results for LLMs.
They suffer from low efficiency and performance.
Others~\cite{alayrac2022flamingo, li2023blip, zhu2023minigpt, liu2023visual, liu2023improved} take an end-to-end approach, first projecting the visual features to the language embeddings, then feeding them to the LLM.
Flamingo~\cite{alayrac2022flamingo} bridges vision-only and language-only models through cross-attention, and trains on multimodal web corpora.
BLIP-2 develops a Query Transformer (Q-Former) to bridge the modality gap and bootstraps vision-language pre-training.
MiniGPT-4~\cite{zhu2023minigpt} utilizes a projection layer to align the visual encoder with LLM.
InstructBLIP~\cite{liu2023improved} utilizes the instruction-aware Q-Former for visual feature extraction.
Notably, LLaVA~\cite{liu2023visual} demonstrates multimodal conversational capabilities via a simple linear layer connecting the visual encoder to the LLM.

\noindent \textbf{Video LLMs}
Incorporating LLMs for video understanding presents more challenges than images.
Recent works mainly focus on constructing conversational video understanding datasets and bridging video features to LLMs through Q-former \cite{zhang2023video,li2023videochat} or a simple linear projection with pooling \cite{maaz2023video}, as illustrated in Fig.~\ref{fig:cmp}.
Video-LLaMA~\cite{zhang2023video} aligns the features of both visual and audio encoders with LLM's embedding space using a video Q-former and an audio Q-former. It is trained on massive video/image-caption pairs and visual-instruction-tuning datasets.
VideoChat~\cite{li2023videochat} utilizes a learnable module to combine video foundation models and LLMs. It also proposes a video-centric instruction dataset, and the model exhibits numerous capabilities such as spatial-temporal reasoning and event localization.
Video-ChatGPT~\cite{maaz2023video} first computes spatial-temporal features of videos, then projects them into LLMs' embedding space via a simple linear layer. This framework is trained on a collected dataset consisting of 100K video-instruction pairs.

\subsection{Evaluation of Video LLMs}
Evaluation of the LLMs~\cite{2023opencompass} and multimodal LLMs~\cite{MMBench,fu2023mme,zhang2023m3exam,shi2023chef,xu2023lvlm,li2023empowering,yu2023mm,li2023seed} reports dozens of metrics across various datasets. Image LLMs have been evaluated on multiple vision-language tasks, such as image captioning, visual question answering, image editing, etc.
However, video LLMs are highly underdeveloped. Current works mainly demonstrate their performance through examples or rely solely on ChatGPT for evaluation without verification.
Video-LLaMA~\cite{zhang2023video} demonstrates two video understanding cases focusing on relevance to sound and visual content, and action recognition ability.
VideoChat~\cite{li2023videochat} emphasizes its descriptive, temporal, and causal ability through examples, and also demonstrates versatile ability through meme explanation, counting, etc.
Video-ChatGPT~\cite{maaz2023video} utilizes GPT-3.5 to evaluate response quality using existing video datasets. However, they did not verify the GPT's ability to assess response quality using the metrics they designed.

\section{Method}

We first review the existing video LLMs in Sec~\ref{sec:revisit}. Then, we present our evaluation pipeline in Sec~\ref{sec:evaluation}. Last, we discuss our Video-LLaVA baseline in Sec~\ref{sec:videollava}.

\subsection{Revisiting Video LLMs}
\label{sec:revisit}

In this section, we look at the interaction between visual (video) and linguistic elements. Existing video LLMs, as shown in Fig.~\ref{fig:cmp} (a-d), all consist of three main components: a visual encoder, LLM, and a video-to-text connector. Video LLMs additionally adapt video features into tokens and add them to the head of the user prompt.

Video LLMs first sample multiple frames and extract visual features using a frozen visual encoder, which is a pre-trained foundation model.
Then, the connector is trained to align the video features with language tokens on video-text pairs in video-language datasets. Existing video LLMs adopts different design of connectors: Video-LLaMA~\cite{zhang2023video} adopts video/audio Q-Former, VideoChat~\cite{li2023videochat} employs Q-Former with Global Multi-Head Relation Aggregator (GMHRA), and Video-ChatGPT~\cite{li2023videochat} uses a simple linear layer with spatial and temporal pooling.
Finally, video LLMs concatenate the adapted visual token with user text prompts and feed into an LLaMA-based \cite{touvron2023llama} LLM for detailed textual response generation.

\subsection{Evaluation}
\label{sec:evaluation}

We employ GPT-based and retrieval-based evaluations to comprehensively assess video LLMs. GPT-based evaluations aim to assess multiple aspects of open-ended responses at a human level. Retrieval-based evaluation, on the other hand, focuses on assessing abilities in downstream applications through action recognition and video text retrieval tasks.

\noindent \textbf{GPT-based evaluation}
While the ability to generate open and diverse response is an impressive and distinguishing feature of LLM-base models, it also makes evaluating video LLMs challenging since the response is open-ended and conversational.
An ideal approach for evaluation is using human feedback, but this method suffers from high labor costs and inconsistent standards.
To overcome these challenges, we use a powerful LLM model GPT-3.5~\cite{OpenAI2023} and design human-validated metrics and prompts to improve the evaluation. We will refer to GPT-3.5 as GPT in the following sections.

\begin{figure}
  \centering
  \includegraphics[width=\linewidth]{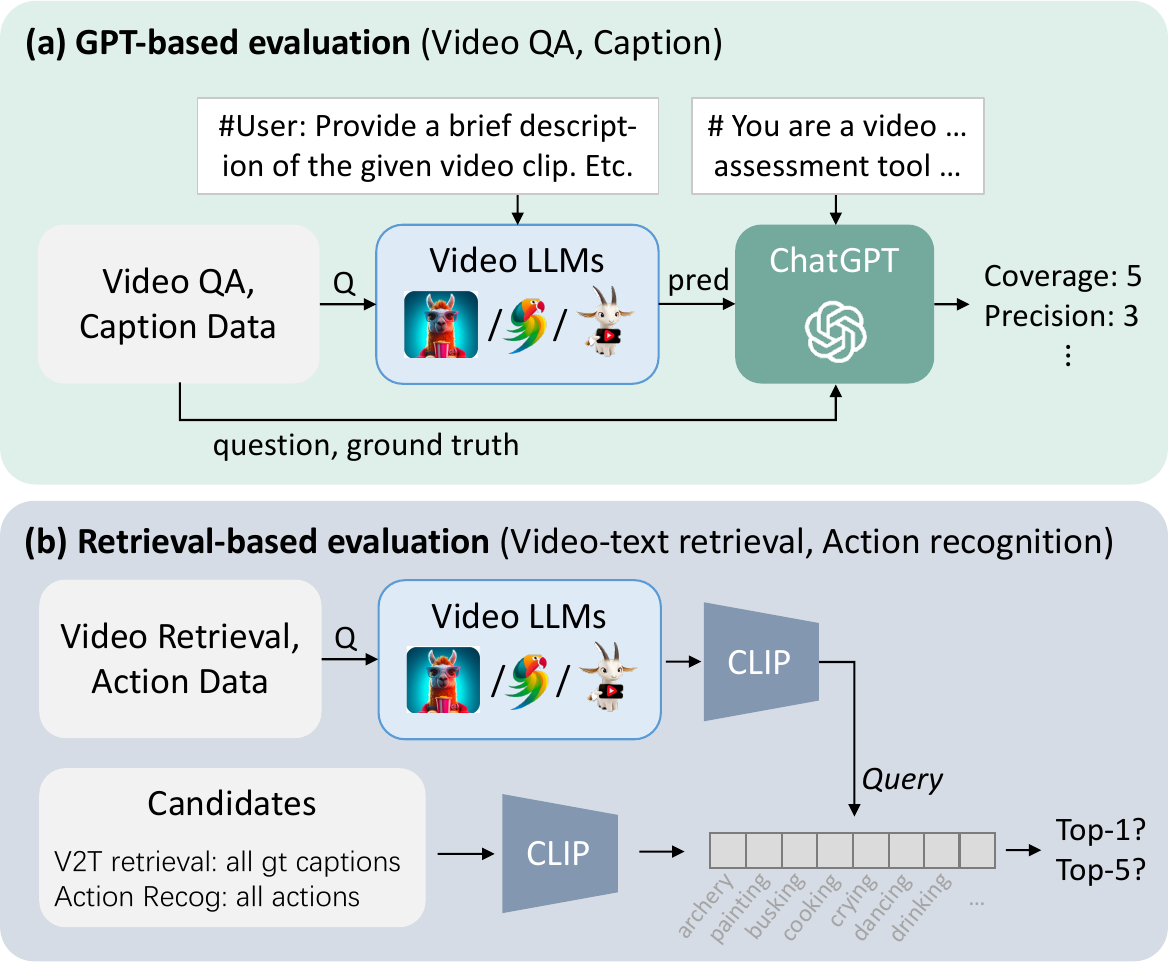}
  \caption{Pipelines of GPT-based and retrieval-based evaluations.}
  \label{fig:GPT_eval}
\end{figure}

Fig.~\ref{fig:GPT_eval} (a) illustrates our GPT-based evaluation pipeline.
GPT scoring is suitable for open-answer tasks such as video question answering (VideoQA) and video captioning.
We build our evaluation on existing VideoQA datasets (MSVD~\cite{wu2017deep}, MSRVTT~\cite{xu2016msr}, TGIF~\cite{li2016tgif}, and ActivityNet-QA~\cite{caba2015activitynet}) as well as MSVD and MSRVTT caption datasets.
During evaluation, we provide GPT with the response from the video LLMs, the correct answer, the task context, and instructions in the prompt. The exact GPT prompts for the evaluations are included in the Supplementary Material. We use the average of all videos as the final score.

We assess the ability to simultaneously understand video and text prompts through the VideoQA task. For this task, we focus on the correctness and degree of matching:
(a) \emph{Correctness}. Open-ended answers require a certain degree of intelligence to determine their correctness. We leverage GPT to understand the question context and provide a true or false judgment for each QA pair.
(b) \emph{Match score}. It is not practical to expect the model to produce an identical response. Since open-ended answers have no clear boundaries of correctness, a match score is necessary to assess the degree of matching between the ground truth answer and the predicted answer. The match score is a relative scale ranging from 1 to 5.

We further assess the ability to accurately understand and describe the video through the video captioning task. Inspired by the widely accepted metrics of recall and precision~\cite{bishop2006pattern}, we propose to evaluate video captions based on coverage and precision scores scaling from 1 to 5.
(a) \emph{Coverage}. High recall is essential for the model to accurately identify the primary content in the video. The coverage score assesses the extent to which the predicted caption contains elements of the ground truth caption.
(b) \emph{Precision}. While having a high recall is desirable, the model must not make redundant guesses. The precision score assesses the extent to which the predicted caption can be verified by the ground truth caption.

Notably, our evaluation penalizes two common failure modes of video LLMs: \emph{verbose output} and \emph{hallucinations}. First, some models produce lengthy responses that contain irrelevant information to the given question or give verbose captions. The match score and precision metrics encourage concise responses by penalizing extra information that is not present in the ground truth. Second, video LLMs can suffer from hallucinations~\cite{liu2023models} and output content that is not present in the original video. This situation cannot be correctly evaluated by traditional n-gram matching evaluations~\cite{liu2023models}.
In our design, hallucinations will be penalized in the precision metric.

\noindent \textbf{Retrieval-based evaluation}
While GPT-based evaluation focuses on open-ended responses, we employ retrieval-based evaluations to assess the capability of VideoLLMs in downstream applications.
Video-text retrieval consists of video-to-text and text-to-video subtasks.
We first use the video LLMs to generate video descriptions, then encode predicted descriptions and ground truth candidates using a CLIP~\cite{radford2021learning} text encoder.
Finally, we use similarity matching for retrieval.
The Text-to-video (T2V) task uses the ground truth text to retrieve the predicted caption, while video-to-text (V2T) uses the predicted caption to retrieve ground truth text.

To evaluate action recognition capability, we perform a retrieval-based evaluation on standard action recognition datasets. As shown in Fig.~\ref{fig:GPT_eval} (b), we query video LLMs for an action label and encode the prediction with the CLIP text encoder. The similarity between the encoded predicted action label and predefined action labels determines action recognition confidence, which is employed to assess action recognition accuracy.

\subsection{Video-LLaVA}
\label{sec:videollava}

As we discussed in \ref{sec:revisit}, the architectural difference between video LLMs mainly lies in the video-to-text connector. To better understand the effect of connector design, we construct a simple baseline using the image LLM LLaVA~\cite{liu2023visual}.

Unlike previous designs that compress video tokens through Q-former or pooling, we adopt a simple approach of feeding all projected visual tokens into the LLM.
The proposed model is named Video-LLaVA, which utilizes pre-trained LLaVA to accelerate the training of videos.
Our model consists of a visual encoder that processes the video input into visual tokens, a linear projector that aligns the different modalities, and an LLM that generates textual responses. This simple design allows for an end-to-end video interaction system.
Fig.~\ref{fig:cmp} (a,e) illustrates our design. Following LLaVA, we adopt CLIP ViT-L/14~\cite{radford2021learning} as the visual encoder and Vicuna-7B~\cite{chiang2023vicuna} as the LLM decoder. We uniformly sample 5 frames, and encode each frame individually.
We directly use the LLaVA linear projector to transform visual tokens and concatenate all visual tokens with language tokens as the input to the LLM.

\begin{table}
  \centering
  \resizebox{\linewidth}{!}{
    \begin{tabular}{l|l|c}
      \toprule
      Dataset                           & Task Domain                             & \# Clips \\
      \midrule
      WebVid~\cite{li2022uniformerv2}   & Video captioning                        & 10M      \\
      NExT-QA~\cite{xiao2021next}       & Video QA                                & 5K       \\
      DiDemo~\cite{rohrbach2015dataset} & Video captioning, temporal localization & 10K      \\
      MSRVTT~\cite{xu2016msr}           & Video QA, video captioning              & 10K      \\
      MSVD~\cite{wu2017deep}            & Video QA, video captioning              & 2K       \\
      TGIF-QA~\cite{li2016tgif}         & Video QA                                & 72K      \\
      HMDB51~\cite{kuehne2011hmdb}      & Action recognition                      & 7K       \\
      UCF101~\cite{soomro2012ucf101}    & Action recognition                      & 13K      \\
      \bottomrule
    \end{tabular}
  }
  \caption{Summary of datasets used in the fine-tuning stage, which are with different kinds of video-related tasks and different lengths. Note that these datasets are used optionally during performance evaluation to maintain a zero-shot setting.}
  \label{tab:sum_dataset}
\end{table}

Tab.~\ref{tab:sum_dataset} presents the datasets used for supervised fine-tuning.
We transform the video-text pairs from different tasks into the unified input sequence template:
\begin{align*}
  \mathtt{User}     & : \mathtt{<Token_{vid}>\ <Token_{ins}>\ <STOP>} \\
  \mathtt{Assistant} & : \mathtt{<Token_{res}>\ <STOP>}
\end{align*}
where $\mathtt{Token_{vid}}$, $\mathtt{Token_{ins}}$, $\mathtt{Token_{res}}$ are video tokens, instruction tokens, and response tokens, respectively.
Since the models and adapter are inherited from LLaVA, we use pre-trained weights from LLaVA and finetune the adapter and LLM decoder using video-instruction pairs.
Specifically, we finetune the model for 10,000 iterations, with a batch size of 64, the AdamW~\cite{loshchilov2017decoupled} optimizer, and a learning rate of 2e-5 with cosine decay.
After training with the above sequences, the model learns to adapt to the video input and generate responses according to the given instructions.

\section{Results}

\newcolumntype{a}{>{\columncolor{cyan!40!blue!60!black!9}}c}
\begin{table}
  \centering
  \small
  \setlength{\tabcolsep}{3.59pt}
  \begin{tabular}{l|ac|ac|ac|ac}
    \toprule
                  & \multicolumn{2}{c|}{V-LLaMA} & \multicolumn{2}{c|}{VideoChat} & \multicolumn{2}{c|}{VChatGPT} & \multicolumn{2}{c}{V-LLaVA} \\
    \midrule
    \it VideoQA   & \it acc     & \it mat        & \it acc     & \it mat          & \it acc     & \it mat         & \it acc     & \it mat       \\
    \:MSVD        & 53.3        & 3.00           & 57.2        & 3.17             & 57.2        & 3.22            & \bf 62.8    & \bf 3.55      \\
    \:MSRVTT      & 24.3        & 1.99           & \bf 46.6    & \bf 2.77         & 42.4        & 2.67            & 41.6        & 2.70          \\
    \:TGIF        & 41.5        & 2.70           & 44.8        & 2.82             & 60.6        & 3.46            & \bf 61.1    & \bf 3.47      \\
    \:A-Net       & 9.8         & 1.33           & 17.8        & 1.74             & 24.5        & 2.01            & \bf 29.5    & \bf 2.19      \\
    \it Average   & 32.2        & 2.26           & 41.6        & 2.63             & 46.2        & 2.84            & \bf 48.8    & \bf 2.98      \\
    \midrule
    \it V-Caption & \it prec    & \it cov        & \it prec    & \it cov          & \it prec    & \it cov         & \it prec    & \it cov       \\
    \:MSVD        & 2.04        & 2.21           & 2.12        & 2.30             & 2.69        & 2.89            & \bf 3.13    & \bf 3.25      \\
    \:MSRVTT      & 1.93        & 1.95           & 1.92        & 2.02             & 2.29        & 2.40            & \bf 2.36    & \bf 2.46      \\
    \it Average   & 1.99        & 2.08           & 2.02        & 2.16             & 2.49        & 2.65            & \bf 2.75    & \bf 2.86      \\
    \midrule
    \it T2V Rtv.  & \it acc$_1$ & \it acc$_5$    & \it acc$_1$ & \it acc$_5$      & \it acc$_1$ & \it acc$_5$     & \it acc$_1$ & \it acc$_5$   \\
    \:MSVD        & 17.8        & 34.6           & 18.7        & 36.3             & 20.4        & 40.4            & \bf 24.8    & \bf  49.9     \\
    \:MSRVTT      & 5.0         & 12.2           & 6.6         & 15.3             & 6.8         & 16.0            & \bf 8.3     & \bf  17.0     \\
    \it Average   & 11.4        & 23.4           & 12.7        & 25.8             & 13.6        & 28.2            & \bf 16.6    & \bf 33.5      \\
    \midrule
    \it V2T Rtv.  & \it acc$_1$ & \it acc$_5$    & \it acc$_1$ & \it acc$_5$      & \it acc$_1$ & \it acc$_5$     & \it acc$_1$ & \it acc$_5$   \\
    \:MSVD        & 17.2        & 35.8           & 23.4        & 44.2             & 24.5        & \bf 49.1        & \bf 27.2    & 48.4          \\
    \:MSRVTT      & 4.3         & 11.0           & \bf 7.3     & 16.4             & 6.4         & \bf 16.5        & 6.9         & 15.3          \\
    \it Average   & 10.8        & 23.4           & 15.4        & 30.3             & 15.5        & \bf 32.8        & \bf 17.1    & 31.9          \\
    \midrule
    \it Act Recog & \it acc$_1$ & \it acc$_5$    & \it acc$_1$ & \it acc$_5$      & \it acc$_1$ & \it acc$_5$     & \it acc$_1$ & \it acc$_5$   \\
    \:K-400       & 21.7        & 37.5           & 30.6        & 49.2             & 33.4        & \bf 56.0        & \bf 34.0    & 55.0          \\
    \:HMDB51      & 20.1        & 44.2           & 27.4        & 53.1             & 33.0        & 61.8            & \bf 40.4    & \bf 62.0      \\
    \:UCF101      & 42.8        & 61.0           & 50.0        & 68.2             & 54.4        & 75.8            & \bf 54.6    & \bf 76.0      \\
    \it Average   & 28.2        & 47.6           & 36.0        & 56.8             & 40.3        & \bf 64.5        & \bf 43.0    & 64.3          \\
    \bottomrule
  \end{tabular}
  \caption{Quantitative comparison of video LLMs on zero-shot video QA, captioning, retrieval, and action recognition. We use the abbreviations V for video, mat for match score, prec for precision, cov for coverage, acc$_1$ and acc$_5$ for top-1 and top-5 accuracy respectively. Evaluation datasets include MSVD~\cite{wu2017deep}, MSRVTT~\cite{xu2016msr}, TGIF~\cite{li2016tgif}, ActivityNet (A-Net)~\cite{caba2015activitynet}, Kinetics-400 (K-400)~\cite{carreira2017quo}, HMDB51~\cite{kuehne2011hmdb} and UCF101~\cite{soomro2012ucf101}.}
  \label{tab:vqa_res_combined}
\end{table}

\begin{figure}
  \centering
  \vspace{-1ex}
  \includegraphics[width=\linewidth]{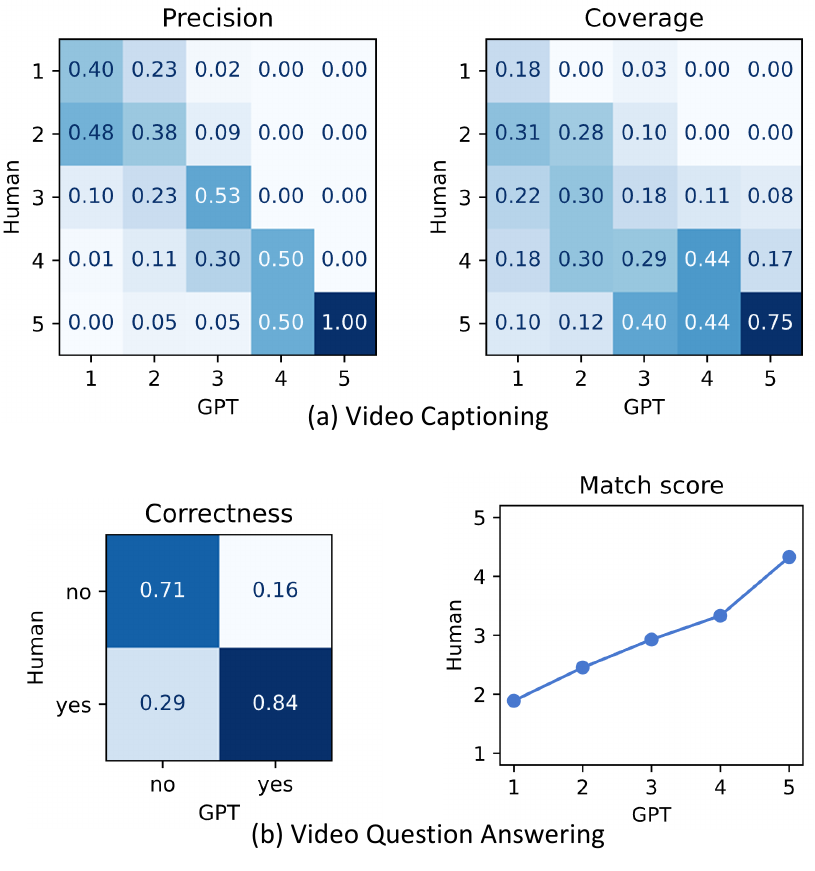}
  \vspace{-2ex}
  \caption{Confusion matrices of GPT scores versus human scores. GPT scores show a high degree of agreement with human scores on the selected criteria.}
  \label{fig:human_eval}
\end{figure}

Video LLMs feature generalization capabilities when handling unseen data, unlike specialized models that primarily utilize abundant supervised data.
To evaluate their zero-shot video understanding ability, we conduct experiments on four video understanding tasks: VideoQA, Video Captioning, Videotext Retrieval, and Action Recognition.
For a fair comparison, we use the 7B versions for all models.

\subsection{Zero-shot Video Question Answering}
In the VideoQA task, we aim to evaluate the model's ability to answer open-ended questions based on a given video and question, which requires fine-grained multimodal understanding.

We evaluate four public open-ended VideoQA datasets: MSVD-QA~\cite{wu2017deep}, MSRVTT-QA~\cite{xu2016msr}, TGIF-QA~\cite{li2016tgif}, and ActivityNet-QA~\cite{caba2015activitynet}.
MSVD-QA and MSRVTT-QA contain five fine-grained types of questions: what, who, how, when, and where.
In TGIF-QA, there are several tasks, including counting objects, identifying an action, recognizing color, localizing, and so on.
ActivityNet-QA features long videos that require obvious spatial-temporal reasoning in QA pairs.

Though different models generate answers in different styles, a human-like GPT assistant is able to assess accurately and flexibly. Utilizing the GPT-based assessment approach, we compare our proposed model with recent models in Tab.~\ref{tab:vqa_res_combined}.
As shown in the table, our Video-LLaVA outperforms previous methods in terms of accuracy and matching score for most datasets. Specifically, these models perform well on short videos, while exhibiting degraded performance on ActivityNet with long videos, indicating the weakness of current models in dealing with long temporal frames.

\begin{figure}
  \centering
  \includegraphics[width=\linewidth]{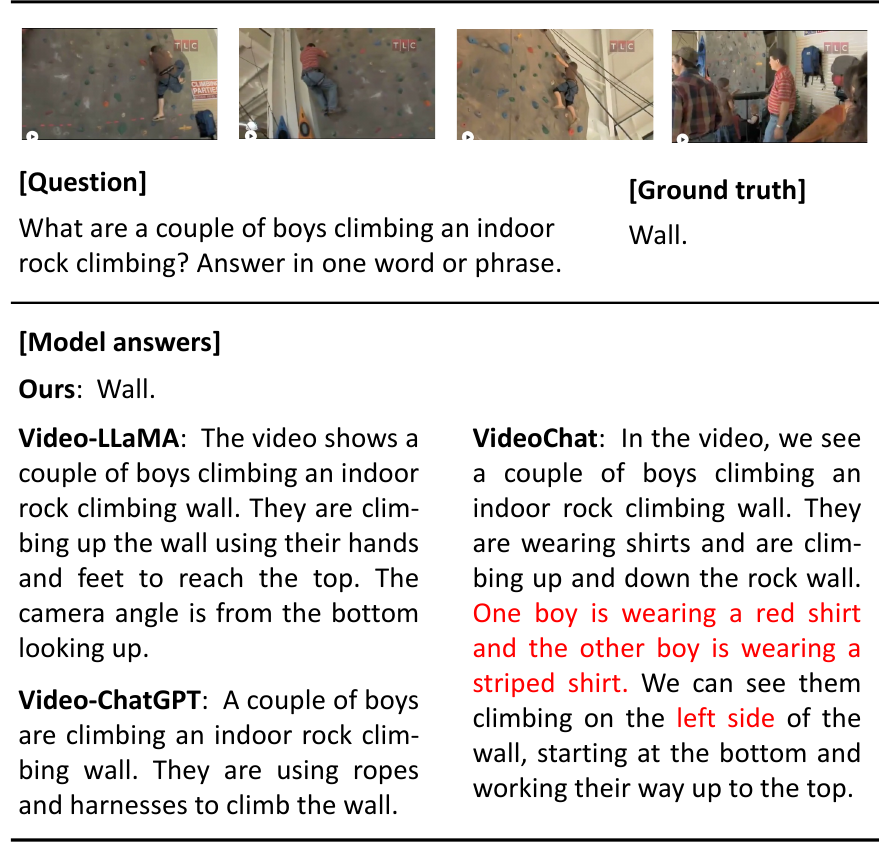}
  \caption{A VideoQA example. We mark factual errors in {\color{red}red}.}
  \label{fig:prompt}
\end{figure}

Among these models, VideoChat usually produces an extremely long answer with overly detailed descriptions, regardless of questions. Video-LLaMA and Video-ChatGPT also fail to produce short responses following the user prompt. On the contrary, our Video-LLaVA is able to answer following the prompt format. In Fig.~\ref{fig:prompt}, we give a typical example to compare the answer modes of different models.
We attribute this ability to crowd-sourced training and formatted prompts~\cite{liu2023improved}.

\subsection{Zero-shot Video Captioning}
Video captioning is a cross-modal open-ended task that generates caption texts to describe the given videos. It is unrealistic to expect a standard answer, as it is possible to describe the video at different levels of granularity.
Therefore, instead of measuring accuracy, we assess caption texts with the precision and coverage metric on a scale of 1-5.
As shown in Tab.~\ref{tab:vqa_res_combined}, our Video-LLaVA achieves the highest precision and coverage compared to other methods, implying that its response is more concise and has fewer hallucinations.

Moreover, we compute the conventional metrics such as CIDEr~\cite{vedantam2015cider}, BLEU-4~\cite{papineni2002bleu}, METEOR~\cite{banerjee2005meteor} and ROUGE-L~\cite{lin2004rouge}, shown in Tab.~\ref{tab:cap_ret_res}. Most methods exhibit very low performance in the zero-shot setting, which also reflects the weakness of these metrics for open-ended captions. On the other hand, our Video-LLaVA achieves the highest performance and outperforms them by a large margin, due to our training in diverse tasks.

\subsection{Zero-shot Video-Text Retrieval}
Video-text retrieval aims to retrieve the matched video or caption from inter-modality candidates. It consists of video-to-text (V2T) and text-to-video (T2V) subtasks. We calculate the text-similarity of generated descriptions and candidates and report Top-1 and Top-5 accuracy metrics in Tab.~\ref{tab:vqa_res_combined}.
Our method outperforms other approaches in the T2V task. In the V2T task, VideoChat, Video-ChatGPT, and Video-LLaVA show comparable performance. However, the relatively low metrics suggest room for future improvement.

\begin{table}
  \centering
  \small
  \setlength{\tabcolsep}{2.5pt}
  \begin{tabular}{l|cccc|cccc}
    \toprule
    \multirow{2}{*}{Method} & \multicolumn{4}{c|}{MSVD-Caption} & \multicolumn{4}{c}{MSRVTT-Caption} \\
                            & C     & B4   & M    & R           & C    & B4   & M    & R             \\
    \midrule
    Video-LLaMA             & 0.0   & 3.7  & 11.3 & 11.5        & 0.0  & 2.8  & 10.5 & 10.6          \\
    VideoChat               & 3.3   & 3.4  & 13.0 & 25.7        & 2.1  & 3.1  & 12.8 & 22.2          \\
    Video-ChatGPT           & 26.2  & 17.4 & 30.2 & 51.0        & 13.7 & 12.4 & 23.9 & 39.8          \\
    Video-LLaVA             & \bf 123.1 & \bf 35.2 & \bf 39.7 & \bf 78.4        & \bf 44.1 & \bf 33.8 & \bf 25.3 & \bf 54.7          \\
    \bottomrule
  \end{tabular}
  \caption{Performance of conventional metrics on video captioning datasets. Higher metric values indicate better results. `C', `B4', `M', and `R' refer to CIDEr~\cite{vedantam2015cider}, BLEU-4~\cite{papineni2002bleu}, METEOR~\cite{banerjee2005meteor} and ROUGE-L~\cite{lin2004rouge}, respectively.}
  \label{tab:cap_ret_res}
\end{table}

\subsection{Zero-shot Action Recognition}
The goal of action recognition tasks is to classify and categorize human actions in videos into a close set of classes. To evaluate the action recognition capability, we use a retrieval-based approach discussed in Sec.~\ref{sec:evaluation}. In Tab.~\ref{tab:vqa_res_combined}, we report the top-1 accuracy and top-5 accuracy on the Kinetics-400~\cite{carreira2017quo}, HMDB51~\cite{kuehne2011hmdb}, and UCF101~\cite{soomro2012ucf101} datasets. Surprisingly, the results show that our simple baseline Video-LLaVA outperforms other counterparts.

\subsection{GPT Scores versus Human Scores}

To validate the GPT evaluation, we collect 200 human feedback samples for each GPT-based metric: precision, coverage, correctness, and match score. We compare these scores with the GPT-rated scores.
As depicted in Fig.~\ref{fig:human_eval}, the confusion matrices demonstrate the strength of the GPT-based evaluation in capturing the performance of video LLMs on selected metrics.
For video captioning evaluation, there is a strong correlation between GPT scores and human scores in terms of precision and coverage. This validates GPT's ability to evaluate video captions.
In VideoQA, since the ground truth answer is typically a single word, it is more difficult for GPT and humans to evaluate. Fig.~\ref{fig:human_eval} (b) shows the confusion matrix of correctness and the average human match scores for given GPT scores. Results show that GPT and humans mostly agree on correctness, and the average human match scores increase monotonically with GPT scores.
The consistent agreement between human and GPT scores verifies the effectiveness of our GPT-based evaluations of video captioning and question answering.


\begin{figure*}
  \centering
  \includegraphics[width=0.98\linewidth]{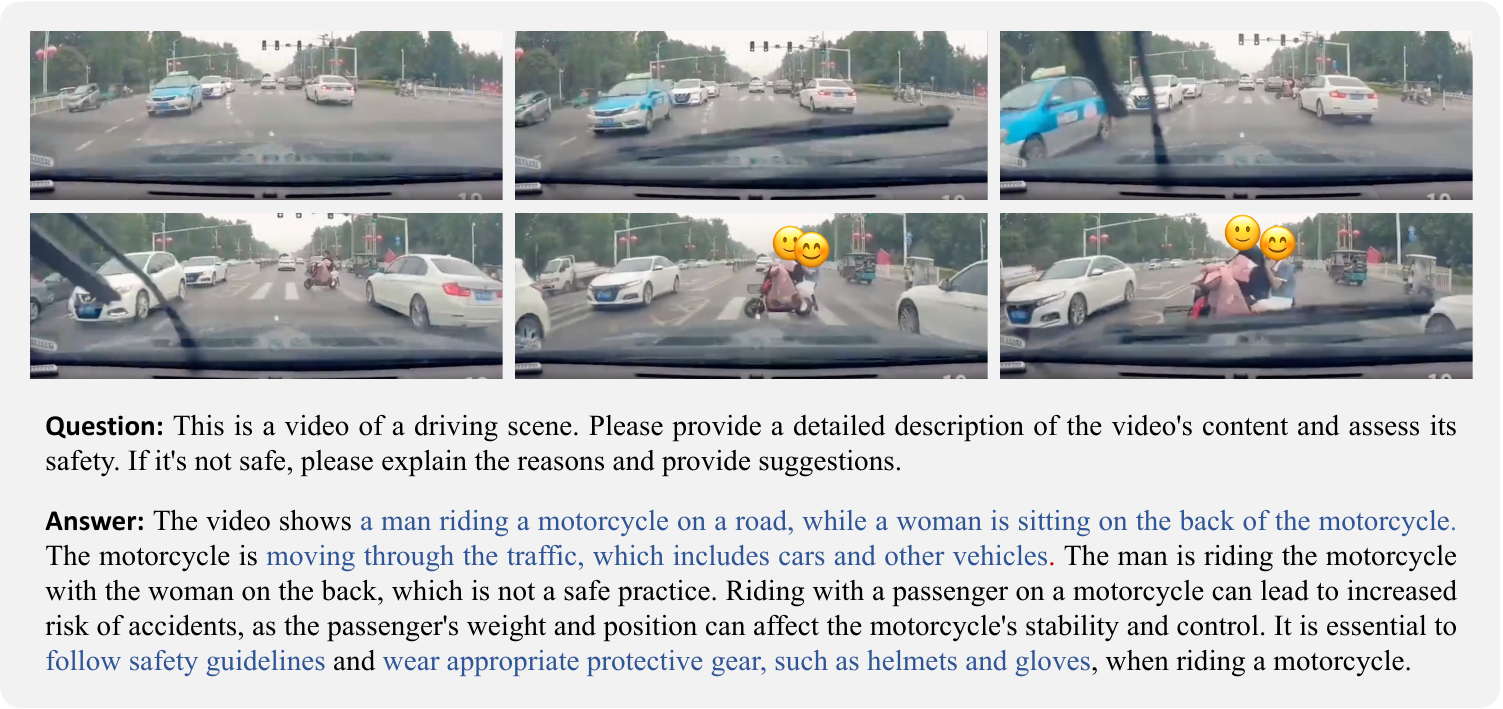}
  \caption{An example response in a dangerous driving scene. Our model reads the dangerous scenario of a motorcycle crossing the road in the video and advises not to carry two people and to wear a helmet.}
  \label{fig:driving_0}
\end{figure*}

\begin{figure*}
  \centering
  \includegraphics[width=0.98\linewidth]{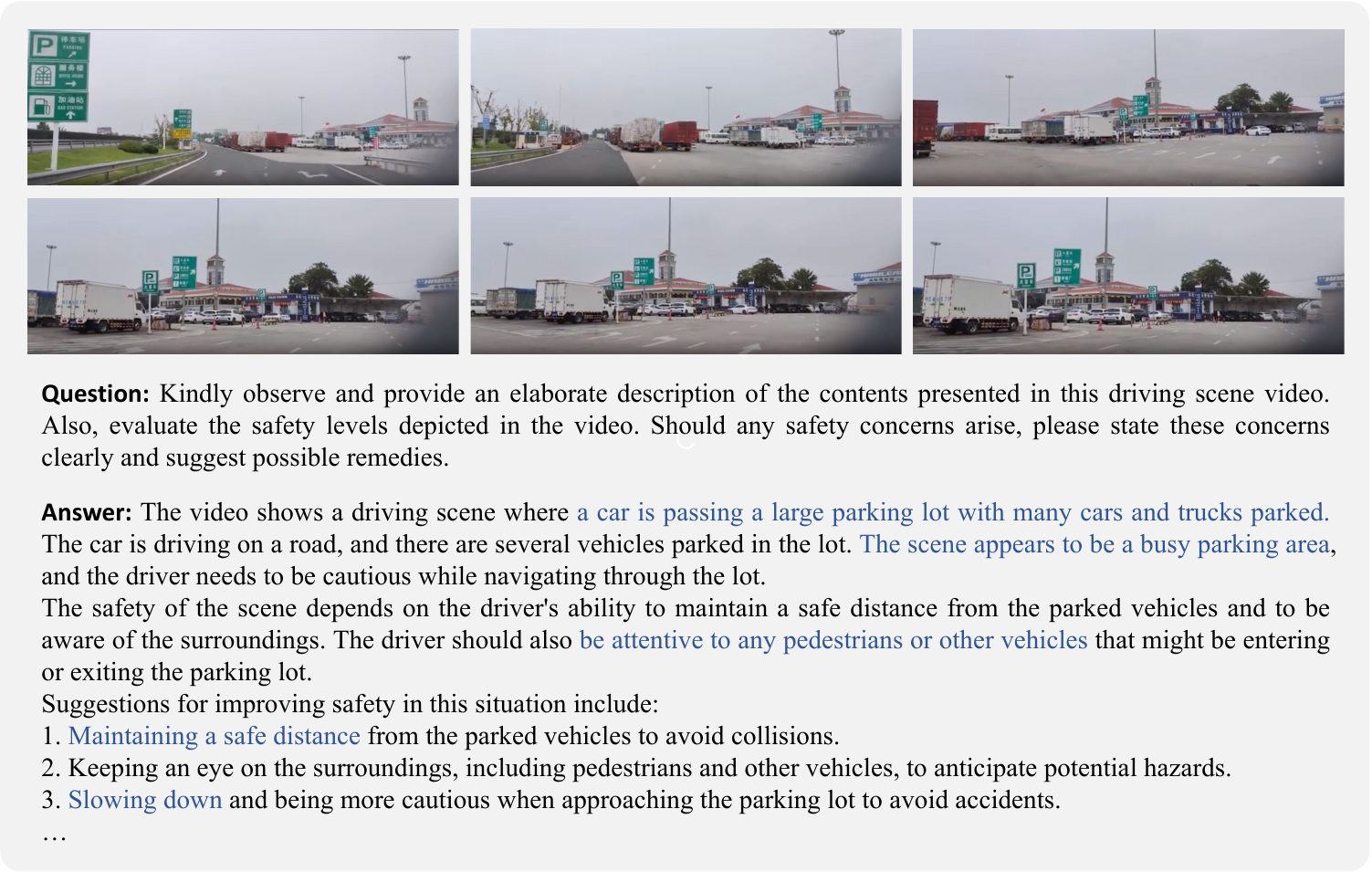}
  \caption{An example response in a normal driving scene. The model reports that the car is passing the parking lot and warns to watch out for pedestrians and other cars and trucks.}
  \label{fig:driving_1}
\end{figure*}

\section{Application: Driving Scene Understanding}

To inspect how video LLMs apply to specific industrial applications beyond academic datasets, we take our Video-LLaVA and driving scenarios as an example and perform an evaluation to exploit the applicability of video LLMs in the real world. Qualitative results verify its diverse capability, as shown in Fig.~\ref{fig:driving_0} and \ref{fig:driving_1}. We observe that our model demonstrates proficiency in open-set object recognition, spatial-temporal modeling, safety reasoning, and practical suggestions.

\textbf{Open-set object recognition.}
Beyond the limitations of traditional closed-set models that can only recognize previously trained objects, open-set object recognition is a pivotal ability for a wide range of pragmatic applications.
Experimental dialogues in Fig.~\ref{fig:driving_0} and \ref{fig:driving_1} show that our model can recognize most vehicles (such as cars, trucks, motorcycles, and bicycles), humans, traffic signs, and roads. Since we only provide a small dataset during the fine-tuning stage, we believe this ability comes from the pre-training stage where the model is trained on abundant open-world datasets.
The ability of open-set recognition can be applied to many tasks, from autonomous driving and industrial robotics to security systems and healthcare diagnostics. In a dynamic and unpredictable world, this potential is significant, making it not only beneficial but also essential.

\textbf{Spatial-temporal modeling.}
Unlike image-based models, spatial-temporal modeling is a core capability of video models. From the keyframes and generated descriptions, we can see that the model exhibits exceptional potential in perceiving and tracking driving scenes. For example, the model says that \textit{the motorcycle is moving through the traffic} and \textit{a car is passing a large parking lot with many cars and trucks parked}.
With its keen perception, the model can accurately detect and interpret complex dynamics within the driving environment, contributing significantly to enhanced safety and predictive decision-making. In essence, our Video-LLaVA leverages general knowledge acquired through large-scale pre-training to understand and reason about the interplay of space and time in driving scenes, thereby offering valuable insights.

\textbf{Safety reasoning and suggestions.}
Besides accurate perception, the model also presents exceptional ability to provide safety reasoning and practical suggestions.
Specifically, it notes the risk when the motorcycle moves through the traffic in Fig.~\ref{fig:driving_0}, and gives concrete advice on following safety guidelines and wearing a protective helmet. In Fig.~\ref{fig:driving_1}, it alerts drivers to watch out for pedestrians and other vehicles entering and exiting the parking lot.
This feature not only enhances the reliability and accuracy of decision-making processes but also significantly contributes to risk mitigation and operational efficiency.

In a nutshell, the Video-LLaVA can be equipped with various capabilities in a unified framework, providing an efficient and comprehensive way for real-world applications. Predictably, this paradigm can also be extended to broader scenarios such as scene prediction and driving planning. It validates the generalization and feasibility of Video-LLaVA in the real world.


\section{Limitation and Future Work}

Despite the promising results of video LLMs, there are several limitations that should be recognized and addressed in future work.

The first is the ability to process long videos. Under the constraints of memory and computation time, VideoLLM models usually try to select video frames or use feature pooling to reduce the computational burden. However, such an approach can hardly adapt to long videos of several minutes due to the loss of intermediate information. For example, we observe in Table~\ref{tab:vqa_res_combined} that the performance of ActivityNet is significantly lower than other datasets with short videos, indicating a large room for model improvement. A promising approach might be to design a memory-based paradigm that allows streaming input and addresses catastrophic forgetting. In this way, both long and short videos can be processed in a unified framework at an acceptable computational cost.

Second, we can only feed frames with small resolutions into the model due to a limited number of tokens. Video with large resolution contains more spatial context, which is significant for real-world scenarios. In the future, VideoLLM models are supposed to be compatible with different scales of inputs to meet the needs of practical tasks, improving their accuracy and adaptability.

Furthermore, most LLM-based models have the phenomenon of hallucination, which means that the model may falsely describe something that does not appear in the videos. The risk of hallucinations comes from the pre-training datasets. For example,  Video-LLaVA occasionally describes a virtual dog on the road for driving scenes. This phenomenon can seriously affect utility and safety, especially for autonomous driving applications.

Given these limitations, future research will focus on optimizing these aspects to improve the model's capacity, speed, accuracy, and generalization ability.

\section{Conclusion}

In this work, we provide a general and comprehensive evaluation of video large language models. A unified GPT-based pipeline is established and verified to assess the open-ended video tasks. Besides, we build a Video-LLaVA model trained on diverse video datasets, achieving SoTA results.
Moreover, through our extensive work, we have broadened the horizons of video LLMs in practical applications, with a particular concentration on driving scene comprehension. By collecting driving videos and meticulous labeling, our model performs well in recognizing real-world objects, reasoning safety, and giving suggestions.
Our work illustrates that the video LLM model can be integrated with versatile capabilities within a unified structure, enabling a highly effective and holistic approach for practical applications.

  {
    \small
    \bibliographystyle{ieeenat_fullname}
    \bibliography{main}
  }


\end{document}